\def\year{2021}\relax
\documentclass[letterpaper]{article} 
\usepackage{sty21}  
\usepackage{times}  
\usepackage{helvet} 
\usepackage{courier}  
\usepackage[hyphens]{url}  
\usepackage{graphicx} 
\usepackage{natbib}  
\usepackage{caption} 
\usepackage{multirow}
\usepackage{graphicx}
\usepackage{amsmath}
\usepackage{amssymb}
\usepackage{xcolor}
\newcommand{\vx}{{\bf x}}
\usepackage[switch]{lineno}

\usepackage{caption}
\usepackage{subcaption}

\title{Few-Shot Lifelong Learning}

\author {

        Pratik Mazumder\thanks{Equal contribution.}\textsuperscript{\rm 1},
        Pravendra Singh$^{\ast}$\textsuperscript{\rm 2},
        Piyush Rai\textsuperscript{\rm 1} \\
}
\affiliations {
    \textsuperscript{\rm 1} Department of Computer Science and Engineering, IIT Kanpur, India\\
    \textsuperscript{\rm 2} Independent Researcher, India \\
    pratikm@cse.iitk.ac.in, pravendra1988@gmail.com, piyush@cse.iitk.ac.in
}

\begin{document}

\maketitle

\begin{abstract}
Many real-world classification problems often have classes with very few labeled training samples. Moreover, all possible classes may not be initially available for training, and may be given incrementally. Deep learning models need to deal with this two-fold problem in order to perform well in real-life situations. In this paper, we propose a novel Few-Shot Lifelong Learning (FSLL) method that enables deep learning models to perform lifelong/continual learning on few-shot data. Our method selects very few parameters from the model for training every new set of classes instead of training the full model. This helps in preventing overfitting. We choose the few parameters from the model in such a way that only the currently unimportant parameters get selected. By keeping the important parameters in the model intact, our approach minimizes catastrophic forgetting. Furthermore, we minimize the cosine similarity between the new and the old class prototypes in order to maximize their separation, thereby improving the classification performance. We also show that integrating our method with self-supervision improves the model performance significantly. We experimentally show that our method significantly outperforms existing methods on the miniImageNet, CIFAR-100, and CUB-200 datasets. Specifically, we outperform the state-of-the-art method by an absolute margin of 19.27\% for the CUB dataset.
\end{abstract}

\section{Introduction}

Deep learning models have successfully matched human beings in many real-world problems. As a result, the number and diversity of applications of deep learning are increasing at a rapid rate. However, deep learning models require training on a large amount of labeled data.  Labeled data is not always available for many real-world problems, and manually labeling data is a costly and time-consuming process. Therefore, recent works have investigated few-shot learning methods \cite{PN,RN,MAML}, which involve a specialized training of models to help them perform well even for classes with very few training samples.

Another common characteristic of real-world problems is that all the training data may not be available initially \cite{ICARL,LWF,EEIL}. New sets of classes may become available incrementally. Therefore, the model has to perform lifelong/continual learning in order to perform well in such settings. The lifelong learning problem generally involves training a model on a sequence of disjoint sets of classes (task) and learning a joint classifier for all the encountered classes. This setting is also known as the class-incremental setting of lifelong learning \cite{EGR,ICARL,EEIL}. Another simpler setting, known as the task-incremental setting, involves learning disjoint classifiers for each task.

In this paper, we proposed a framework for the few-shot class-incremental learning (FSCIL) problem. The incremental nature of training makes the few-shot learning (FSL) problem even more challenging. On the other hand, humans can continuously learn new categories from very few samples of data. Therefore, to achieve human-like intelligence, we need to equip deep learning models with the ability to deal with the few-shot class-incremental learning problem. 

Training the entire network on classes with very few samples will result in overfitting, which will hamper the network's performance on test data. Additionally, since the model will not have access to old classes when new classes become available for training, the model will suffer from catastrophic forgetting \cite{CF} of the old classes. Therefore, in order to solve the FSCIL problem, we have to address the two issues of overfitting and catastrophic forgetting simultaneuously, which makes it even harder.

A common approach of preventing catastrophic forgetting is to ensure that, while training on new classes, the model's output logits corresponding to the older classes remain unchanged. To achieve this, many methods \cite{ICARL,PDR,NCM,EEIL} use a knowledge distillation loss \cite{KD}. The distillation loss can be computed on a few ``replay"-ed samples from the old classes. However, the distillation loss is generally biased towards classes with more samples and the new classes. Recently, the authors in \cite{tao2020few} proposed a method TOPIC to solve the few-shot class-incremental learning problem, using a cognition-based knowledge representation technique. TOPIC uses a neural gas (NG) network~\cite{NG,GNG} to model the topology of feature space. While training on new classes, it keeps the topology of NG stable and pushes the samples of new classes to their respective NG node to preserve old knowledge.

We propose a novel method, called few-shot lifelong learning (FSLL) for the FSCIL problem by addressing the overfitting and catastrophic forgetting problems from the perspective of the trainable parameters. When a new set of classes becomes available for training, we do not train the entire model on it since the new classes have very few examples, and the full model will quickly overfit to these examples. Instead, we choose very few session trainable parameters to train on these new classes, which reduces the overfitting problem. Our method selects these session trainable parameters in such a way that only unimportant parameters of the model get chosen. As a result, the training on the new set of classes only affects a few unimportant model parameters. Since the important parameters in the model are not affected, the model can retain the old knowledge, thereby minimizing catastrophic forgetting. We encourage the session trainable parameters to be properly updated, but not deviate far from their previous values. To ensure this, we add a regularization loss on the session trainable parameters. Additionally, we maximize the separation between the new and the old class prototypes by minimizing their cosine similarity to improve the network classification performance. We also explore a variant of our method that uses self-supervision as an auxiliary task to improve the model performance further.

We perform experiments on the miniImageNet \cite{miniImageNet}, CIFAR-100 \cite{cifar}, and CUB-200~\cite{cub200} datasets in the FSCIL setting and compare our performance with state-of-the-art method and other baselines. Our experimental results show the effectiveness of our method. We also perform extensive ablation experiments to validate the components of our method. Our main contributions are as follows:

\begin{itemize}
\item We propose a novel method for the few-shot class-incremental learning problem. Our proposed method selects very few unimportant model parameters to train every new set of classes in order to minimize overfitting and the catastrophic forgetting problem. 
\item We empirically show that using self-supervision as an auxiliary task can further improve the performance of the model in this setting.
\item We experimentally show that our proposed method significantly outperforms all baselines and the state-of-the-art methods for all the compared datasets.
\end{itemize}

\section{Proposed Method}
\subsection{Problem Setting}

In the FSCIL setting , we have a sequence of labeled training sets $D^{(1)}, D^{(2)}, \cdots$, where $D^{(t)} = \lbrace (\vx_j^{(t)}, y_j^{(t)}) \rbrace_{j=1}^{\vert D^{(t)} \vert}$. $L^{(t)}$ represents the set of classes of the $t^{th}$ training set, where $\forall i,j,\ L^{(i)} \cap L^{(j)}=\varnothing$ for $i\neq j$. The first training set, $D^{(1)}$, consist of base classes with a reasonably large number of training examples per class. The remaining training sets $D^{(t>1)}$ are few-shot training sets of new classes, where each class has very few training samples. The model has to be incrementally trained on $D^{(1)}, D^{(2)}, \cdots$ and only $D^{(t)}$ is available at the $t^{th}$ training session. After training on $D^{(t)}$, the model is evaluated on all the encountered classes in $L^{(1)}, \cdots, L^{(t)}$. Each of the few-shot training sets $D^{(t>1)}$, contain $C$ classes and $K$ training examples per class. This setting is referred a $C$-way $K$-shot FSCIL setting. 

Since every few-shot training set has very few examples per class, storing examples from such classes will effectively violate the incremental learning setting. Therefore, our proposed method does not store any examples from the previously seen classes. Also, since FSCIL is based on the class-incremental setting, no task information is available during test time, and the model has to perform classification jointly for all the seen classes.

\begin{figure}[t]
\begin{center}
    \includegraphics[width=0.40\textwidth]{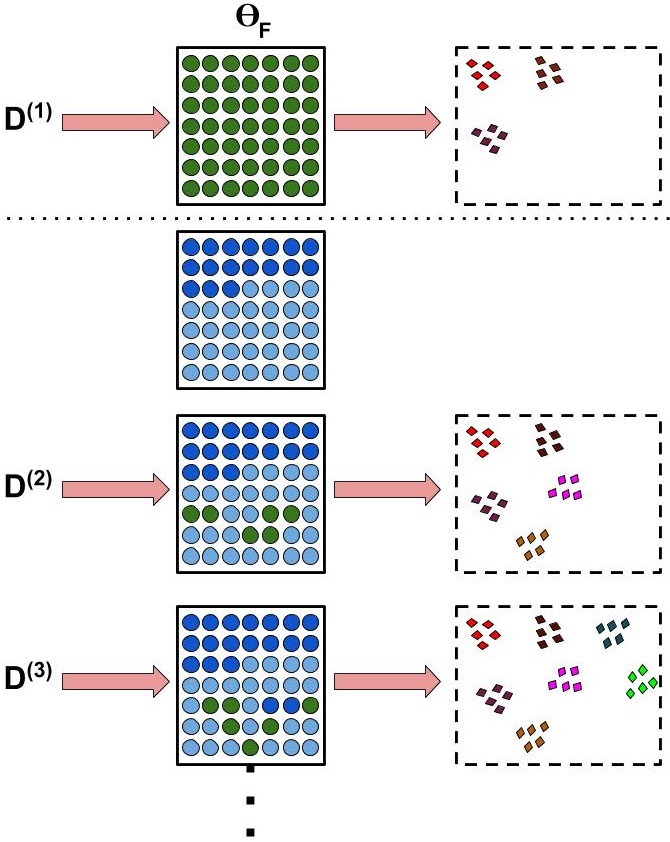}
\end{center}
\caption{\textbf{Our proposed few-shot lifelong learning method}. We initially train the network on the base training set $D^{(1)}$ that contains many examples per class. In the first session, all the parameters are trainable (marked in green). After finishing the training on $D^{(1)}$, we find the important (marked in deep blue) and unimportant parameters (marked in light blue) in the model. For training on the few-shot training set $D^{(t>1)}$, we select a few unimportant parameters as the session trainable parameters (marked in green). After completing the training on each few-shot training set $D^{(t>1)}$, we re-identify the important and unimportant parameters and select the few session trainable parameters for the next session. By preserving the important parameters in the model, the model can preserve the old knowledge. Further, by training only a few session trainable parameters for each few-shot training set, overfitting is also reduced.} 
\label{fig:method}
\end{figure}

\subsection{Method Overview}
In the FSCIL setting, the naive approach is to incrementally train on the training sets $D^{(1)}, D^{(2)}, \cdots$. However, this approach will lead to a catastrophic forgetting of the older classes. Additionally, since the training set $D^{(t>1)}$ has very few training examples per class, the model will overfit to these few examples. We propose a novel method to deal with these two challenges.

Our network consists of a feature extractor $\Theta_F$ and a fully connected classifier $\Theta_C$. In our approach, we first train the complete network on the base training set $D^{(1)}$ for classification using the cross-entropy loss similar to TOPIC. This is a common practice in the few-shot learning setting. During this session, all the parameters of the network are trainable (Fig. \ref{fig:method}).
\begin{equation}
    L_{D^{(1)}}(\vx,y) = F_{CE}(\Theta_C(\Theta_F(\vx)),y)
\end{equation}
where $\vx$ and $y$ refer to an image and its label, $(\vx,y) \in D^{(1)}$, $F_{CE}$ refers to the cross entropy loss. 

We discard $\Theta_C$ after completing the training on $D^{(1)}$. Using the trained feature extractor $\Theta_F$, we extract the features of all the training samples of $D^{(1)}$ and average them class-wise to obtain the class prototypes of the base classes. 
\begin{equation}\label{eq:proto}
    Pr[c] = \frac{1}{N_c} \sum\limits_{k=1}^N  \mathbb{I}_{(y_k = c)} (\Theta_F(\vx_k))
\end{equation}
where $Pr[c]$ is the prototype of class $c$ in $D^{(t)}$, $N_c$ is the number of training examples in the class $c$, $N$ is the number of training examples in $D^{(t)}$, $\forall k$ $(\vx_k,y_k) \in D^{(t)}$. $\mathbb{I}_{(y_k = c)}$ is an indicator function that returns 1 only when the label of the sample $x_k$ is the class $c$, otherwise it returns 0. 

The session 1 test examples belong to all the classes encountered till now, i.e., all the base training set classes. For each test example, in session 1, we find the nearest class prototype and predict that class as the output class.

When the session $t>1$ starts, the $D^{(t>1)}$ training set becomes available and data from all previous training sets $\{D^{(1)},D^{(2)},\cdots D^{(t-1)}\}$ become inaccessible. We select very few unimportant parameters from $\Theta_F$ for training on the few-shot classes in session $t$ (Fig. \ref{fig:method}). We refer to these parameters as the session trainable parameters $P^t_{ST}$ for session $t$. Parameters/weights with low absolute magnitude contribute very less to the final accuracy and are, therefore, unimportant \cite{han2015learning}. In order to select the session trainable parameters, we choose a threshold for each layer in $\Theta_F$. All parameters in a layer having absolute value lower than the threshold are chosen as the session trainable parameters $P^t_{ST}$ for session $t$.  We provide an ablation in Sec. \ref{sec:threshold} to show the effect of the proportion of session trainable parameters on the final accuracy. Since we choose the parameters with the lowest absolute weight values, it is highly unlikely that the ``high" importance/absolute weight valued parameter will get selected as a less important parameter for the subsequent tasks.

We refer to the remaining parameters as the knowledge retention parameters $P^t_{KR}$ for session $t$, and we keep them frozen during this session. Since we choose only the unimportant parameters for training, the important parameters remain intact in the model. Therefore, our approach prevents the loss of knowledge from the previously seen classes and reduces catastrophic forgetting. 

We train the session trainable parameters $P^t_{ST}$ on a triplet loss, in order to bring examples from the same class closer and push away those from different classes. 
\begin{multline}
    L_{TL}(x_i,x_j,x_k) = max(d(\Theta_F(x_i), \Theta_F(x_j))- \\d(\Theta_F(x_i), \Theta_F(x_k)),0)
\end{multline}
where $x_i,x_j,x_k$ are the images in $D^{(t>1)}$, $L_{TL}$ refers to the triplet loss, $d$ refers to euclidean distance. Let $y_i,y_j,y_k$ be the corresponding class labels of $x_i,x_j,x_k$ and  $y_i=y_j,y_i \neq y_k$.

We encourage the session trainable parameters to be properly updated, but not deviate far from their previous values. We apply a regularization loss on $P^t_{ST}$ to achieve this goal. For the regularization loss, we use $\ell_1$-regularization between the current $P^t_{ST}$ parameters weights and their previous values.
\begin{equation}
    L_{RL} = \sum\limits_{i= 1}^{N_p^t} ||w_i^{t}-w_i^{t-1}||_1
\end{equation}
where $N_p^t$ refers to the number of session trainable parameters $P^t_{ST}$ for the training set $t$. $w_i^{t},w_i^{t-1}$ refer to the current and previous weights of the $i^{th}$ parameter in $P^t_{ST}$.

Additionally, we apply a cosine similarity loss to minimize the similarity between the prototypes of the older classes $Pr^{prev}$ and those of the new classes $Pr^t$. The new class prototypes are computed using Eq. \ref{eq:proto} for $D^{(t)}$. 
\begin{equation}
    L_{CL} = \sum\limits_{i=1}^{N_{Pr}^t}  \sum\limits_{j=1}^{N_{Pr}^{prev}}F_{cos}(Pr^t[i],Pr^{prev}[j])
\end{equation}
where $Pr^t$ refers to the prototypes of $D^{(t)}$, $Pr^{prev}$ refers to set of prototypes from all the previous classes. $N_{Pr}^t,N_{Pr}^{prev}$ refer to the number of class prototypes in the current training set and all the previous training sets respectively. $F_{cos}$ refers to the cosine distance loss. $Pr^t[i]$ $Pr^{prev}[j]$ refer to the $i^{th}$ and $j^{th}$ prototypes in $Pr^t$ and $Pr^{prev}$, respectively.

Therefore, the total loss for the training set $D^{(t>1)}$ is as follows:
\begin{equation}
L(D^{(t>1)}) = L_{TL} + L_{CL} + \lambda L_{RL}  
\end{equation}
where $\lambda$ is a hyper-parameter that determines the contribution of the regularization loss.

 After completing the training on $D^{(t>1)}$, we extract the features of the training samples of all the classes in the current training set using the trained feature extractor $\Theta_F$ and compute the class-wise mean/prototype of these features (Eq. \ref{eq:proto}). 
 We perform the nearest prototype-based classification using the prototypes of all classes to predict the nearest class for each test example in the current session.

\subsection{Self-Supervised Auxiliary Task}
We also experiment with a variant of our method, where we train the complete network on $D^{(1)}$ using the standard cross-entropy loss and an auxiliary self-supervision loss. We use rotation prediction as our auxiliary task \cite{gidaris2018unsupervised}. 

In order to add the auxiliary task, we add a rotation prediction network $\Theta_R$ after $\Theta_F$, in parallel with $\Theta_C$. We rotate the each training sample in $D^{(1)}$ by either 0, 90, 180, and 270 degrees and train the network to predict the angle of rotation using $\Theta_R$. The image feature extracted by $\Theta_F$ is given to $\Theta_R$ for the rotation prediction task. The total loss for training on $D^{(1)}$ in this case is as follows:

\begin{multline}
    L_{D^{(1)}}(\vx,y) = F_{CE}(\Theta_C(\Theta_F(\vx)),y) +\\ F_{CE}(\Theta_R(\Theta_F(\vx)),y^r)
\end{multline}
where  $(\vx,y) \in D^{(1)}$ and $y^r$ is the angle of rotation that $\vx$ was rotated by.

We empirically show that the performance of our method can be improved further using self-supervision. In the ablation studies Sec. \ref{sec:auxtask}, we experimentally show that the rotation prediction-based self-supervision task performs better than SimClr and patch location prediction methods when used as an auxiliary task in our method.

\section{Related Work}
The lifelong/continual learning problem can have two settings: class-incremental setting and task-incremental setting. 

\subsection{Class-Incremental Lifelong Learning}
Class-incremental lifelong learning involves training a model on multiple sets of disjoint classes in a sequence and testing on all the encountered classes. iCaRL, is a popular method proposed in \cite{ICARL}, which stores class exemplars and learns using the nearest neighbor classification loss on the new classes and a distillation loss on the old class exemplars. The work in \cite{EEIL} proposes EEIL, which trains the model using a cross-entropy loss and a distillation loss. NCM \cite{NCM} uses cosine distance metric to reduce the bias of the model towards the new classes. Similarly, BIC \cite{BIC} learns a correction model to reduce the bias in the output logits. We focus on class-incremental learning but in a few-shot setting, which is a more challenging problem due to the few-shot nature of the classes.

\subsection{Task-Incremental Lifelong Learning}

Task-incremental lifelong learning involves training a model on multiple tasks (which are disjoint sets of classes) in a sequence but maintaining a separate classifier for each task. As a result of a reduced search space, this setting is simpler than the class-incremental setting. Task-incremental lifelong learning methods can be of three types: a) regularization-based, b) replay-based, and c) dynamic network-based.

Regularization-based methods try to reduce changes in the output logits/ important parameters of the network while training on new tasks in order to preserve the old task knowledge \cite{IMM, SI,REWC}. The work in \cite{LWF} uses knowledge distillation to achieve this goal. EWC \cite{EWC} decreases the learning rate for the parameters that are important to the older tasks.

Replay-based methods \cite{GEM,AGEM}, store exemplars from old tasks and include them in the training process of the new tasks in order to reduce catastrophic forgetting. Some methods utilize generative models to generate data for the old tasks instead of storing the exemplars \cite{DGR,MGR,LGAN,xiang2019incremental}. 

Dynamic network-based methods modify the network to train on new tasks \cite{PackNet,Piggyback,MAS,HAT,DEN}. These methods employ techniques such as dynamic expansion, network pruning, and parameter masking to prevent catastrophic forgetting. PackNet, a method proposed in \cite{PackNet}, utilizes pruning to free parameters for training new tasks. The work in \cite{HAT} proposes to learn attention masks for old tasks to constrain the parameters when training on the new task. The authors in \cite{xu2018reinforced} utilize reinforcement learning to decide the number of additional neurons needed for each new task.

Since we focus on the class incremental setting in this paper, the task-incremental methods do not apply to this setting. Therefore, we have to exclude them for comparison in the experimental section.

\subsection{Few-Shot Learning}

Few-shot learning (FSL) methods train models to perform well for classes with very few training examples (few-shot classes). Many research works deal with the few-shot learning problem. Few-shot learning methods generally employ meta-learning and metric learning techniques \cite{miniImageNet,PN,RN,MAML,MTL}. However, most of them are focused solely on the few-shot classes. Recently some methods have also explored the loss of performance in the non few-shot classes due to the techniques used to benefit the few-shot classes \cite{DFS,IFS}. The method proposed in \cite{DFS} extends an object recognition system with an attention-based few-shot classification weight generator and redesigns the classifier as a similarity function between feature representations and classification weight vectors. It combines the recognition of both the few-shot and non few-shot classes. 

Most of the standard few-shot learning methods perform testing on few-shot episodes, containing a few classes with very few labeled samples. By reducing the search space to the few classes present in the episode, the problem becomes much simpler. On the other hand, the few-shot class-incremental learning setting performs testing on all the encountered classes, which is more realistic and challenging. TOPIC, proposed in \cite{tao2020few}, utilizes a neural gas network~\cite{NG,GNG} to model the topology of feature space and stabilizes the topology while introducing new classes in order to preserve old knowledge. This method achieves state-of-the-art results in the FSCIL setting, and we have compared our results with this method.
     
\subsection{Self-Supervised Learning}

While obtaining labeled data is expensive and time-consuming, recent work has considered alternative mechanisms that can substitute for such explicitly labeled supervision. In particular, the self-supervised learning paradigm trains a network on data using labels extracted from the data itself. Self-supervised learning helps the network to learn better features. To perform self-supervised learning, various types of pseudo tasks/labels are used to train the network, such as image inpainting or image completion \cite{pathak2016context}, image colorization  \cite{larsson2016learning,zhang2016colorful}, prediction of relative patch position \cite{doersch2015unsupervised}, solving   Jigsaw puzzles in \cite{noroozi2016unsupervised}. In \cite{gidaris2018unsupervised}, the authors propose to rotate the images by a fixed set of angles, and the network is trained to predict the angle of rotation. It is a very popular method for self-supervision.
  
Contrastive Multiview Coding (CMC) \cite{tian2019contrastive} trains the network to maximizes the mutual information between the different views of an image but requires a specialized architecture, including separate encoders for different views of the data. Momentum Contrast (MoCo) \cite{he2020momentum} matches encoded queries q to a dictionary of encoded keys using a contrastive loss, but it requires a memory bank to store the dictionary. SimCLR \cite{chen2020simple} augments the input to produce two different but correlated views and uses contrastive loss to bring them closer in the feature space. It does not require specialized architectures or a memory bank and still achieves state-of-the-art unsupervised learning results, outperforming CMC and MoCo. 

\section{Experiment}

\subsection{Datasets}
We perform experiments in the FSCIL setting using three image classification datasets CIFAR-100~\cite{cifar}, miniImageNet~\cite{miniImageNet} and CUB-200~\cite{cub200}. The CIFAR-100 dataset consists of 100 classes with each class containing 500 training images and 100 testing images. Each of the 60,000 images is of size $32 \times 32$. The miniImageNet dataset also consists of 60,000 images from 100 classes, chosen from the ImageNet-1k dataset \cite{ImageNet}. There are 500 training and 100 test images of size $84\times 84$ for each class. The CUB-200 dataset consists of about 6,000 training images and 6,000 test images for 200 categories of birds. The images are resized to $256\times 256$ and then cropped to $224\times 224$ for training.  

In the case of the CIFAR-100 and miniImageNet datasets, we choose 60 and 40 classes as the base and new classes, respectively. For every few-shot training set, we use a 5-way 5-shot setting, i.e., each few-shot training set has 5 classes with 5 training examples per class. Therefore, we have 1 base training set and 8 few-shot training sets (total 9 training sessions) for the CIFAR-100 and miniImageNet datasets. For the CUB-200 dataset, we choose 100 and 100 classes as the base and new classes, respectively. For every few-shot training set of CUB-200, we use a 10-way 5-shot setting, i.e., each few-shot training set has 10 classes with 5 training examples per class. Therefore, we have 1 base training set and 10 few-shot training sets (total 11 training sessions) for the CUB-200 dataset. We construct each few-shot training set by randomly choosing 5 training examples per class, while the test set contains test examples from all the encountered classes. For a fair comparison, we used the same dataset settings as used in \cite{tao2020few}.

\subsection{Implementation Details}
We use ResNet-18 \cite{ResNet} architecture for our experiments on all the three datasets. The last classification layer of ResNet-18 is $\Theta_C$, and the remaining network serves as the feature extraction network $\Theta_F$. We train $\Theta^{(1)}_F$, and $\Theta^{(1)}_C$ on the base training set $D^{(1)}$ with an initial learning rate of 0.1 and mini-batch size of 128. After the 30 and 40 epochs, we reduce the learning rate to 0.01 and 0.001, respectively. We train on $D^{(1)}$ for a total of 50 epochs. After completing the training on the base training set, we discard $\Theta^{(1)}_C$.

We finetune the feature extractor on each of the few-shot training sets $D^{(t>1)}$ for 30 epochs, with a learning rate of $1e\texttt{-}4$ (and $1e\texttt{-}3$ for CUB-200). We set the threshold values for each layer in such a way that only 10\% of $\Theta_F$ get selected as the session trainable parameters in all our experiments. Since the few-shot training sets contain very few training examples, the mini-batch contains all the examples. After training the feature extractor on $D^{(t)}$, we test $\Theta_F^{t}$ on the combined test sets of all encountered classes. Session $t$ accuracy refers to the total accuracy over all the the classes encountered till that session  ($L^{(1)},L^{(2)} \cdots, L^{(t)}$).

We perform standard random cropping and flipping for data augmentation proposed in \cite{ResNet,NCM} for all methods. Since we have very few new class training samples, we use the batchnorm statistics computed on $D^{(1)}$ and fix the batchnorm layers while finetuning on $D^{(t>1)}$ as done in \cite{tao2020few}. We run each experiment 10 times and report the average test accuracy over all the encountered classes. The standard deviations among the runs are low (around 0.5\% on average) for all the experiments. For the experiments with the auxiliary self-supervision task, we use a convolutional neural network as $\Theta_R$ to predict the rotation angle. $\Theta_R$ consists of 4 convolutional layers, each containing 512 filters of filter size of 3, stride 1, and padding 1. We use an adaptive average pooling layer and a linear layer of output size 4 after the last convolutional layer. For reporting the results, we also include the base set $D^{(1)}$ accuracy for a fair comparison with TOPIC and other methods. 

\begin{table*}[t]
\footnotesize
\centering
\caption{Results on CUB-200 using the ResNet-18 architecture on the 10-way 5-shot FSCIL setting. We compare our method with TOPIC (CVPR'20) which is the state-of-the-art method for this setting. Session $t$ accuracy refers to the total accuracy over all the the classes encountered till that session  ($L^{(1)},L^{(2)} \cdots, L^{(t)}$).}
\resizebox{\textwidth}{!}{
\begin{tabular}{lcccccccccccc}
\hline
\multirow{2}{*}{Method} & \multicolumn{11}{c}{Sessions} & Our Relative \\
  \cline{2-12} & 1 & 2 & 3 & 4 & 5 & 6 & 7 & 8 & 9 & 10 & 11 & Improvements \\
  \hline

Ft-CNN~\cite{tao2020few} & 68.68 & 44.81 & 32.26 & 25.83 & 25.62 & 25.22 & 20.84 & 16.77 &  18.82 & 18.25 &  17.18 & \textbf{+28.37} \\
Joint-CNN~\cite{tao2020few} & 68.68 & 62.43 & 57.23 & 52.80 & 49.50 & 46.10 & 42.80 & 40.10 & 38.70 & 37.10 & 35.60 & \textbf{+9.95} \\ 
iCaRL~\cite{ICARL} & 68.68 & 52.65 & 48.61 & 44.16 & 36.62 & 29.52 & 27.83 & 26.26 & 24.01 & 23.89 & 21.16 & \textbf{+24.39} \\
EEIL~\cite{EEIL} & 68.68 & 53.63 & 47.91 & 44.20 & 36.30 & 27.46 & 25.93 & 24.70 & 23.95 & 24.13 & 22.11 & \textbf{+23.44} \\
NCM~\cite{NCM} & 68.68 & 57.12 & 44.21 & 28.78 & 26.71 & 25.66 & 24.62 & 21.52 & 20.12 & 20.06 & 19.87 & \textbf{+25.68} \\
TOPIC~\cite{tao2020few} & 68.68 & 62.49 & 54.81 & 49.99 & 45.25 & 41.40 & 38.35 & 35.36 & 32.22 & 28.31 & 26.28 & \textbf{+19.27} \\
\textbf{FSLL (Ours)} & \textbf{68.72} & \textbf{65.67} & \textbf{62.33} & \textbf{58.10} & \textbf{55.44} & \textbf{52.66} & \textbf{51.17} & \textbf{50.27} & \textbf{48.31} & \textbf{47.25} & \textbf{45.55} & \textbf{0}\\
\hline
\textbf{FSLL* (Ours)} & \textbf{72.77} & \textbf{69.33}& \textbf{65.51}& \textbf{62.66}& \textbf{61.10}& \textbf{58.65}& \textbf{57.78}& \textbf{57.26}&\textbf{55.59}& \textbf{55.39}& \textbf{54.21} & - \\

\textbf{FSLL*+SS (Ours)} & \textbf{75.63} &  \textbf{71.81} &  \textbf{68.16} &  \textbf{64.32} &  \textbf{62.61} &  \textbf{60.10} &  \textbf{58.82} &  \textbf{58.70} & \textbf{56.45} &  \textbf{56.41} &  \textbf{55.82} & -\\

\hline

\end{tabular}
}
\label{tab:CUB-200}
\end{table*}
\subsection{Baselines and Compared Methods}

We compare our method with iCARL \cite{ICARL}, EEIL~\cite{EEIL} and NCM~\cite{NCM} in the FSCIL setting as in \cite{tao2020few}. We compare our method with the Ft-CNN, which involves only finetuning the model on the few training examples of $D^{(t>1)}$. We also compare our method with the Joint-CNN method, which trains on the combined data of the base and few-shot classes.

\begin{figure}[t]
\begin{center}
    \includegraphics[width=0.45\textwidth]{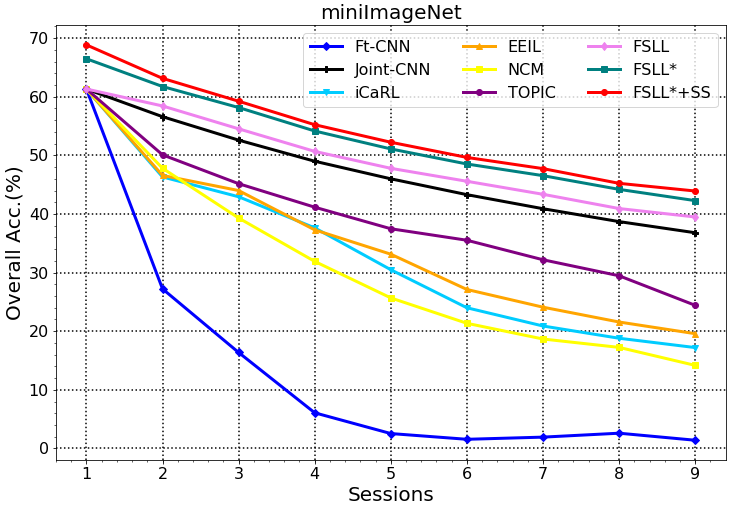}
\end{center}
\caption{Results on miniImageNet using the ResNet-18 architecture on the 5-way 5-shot FSCIL setting}
\label{fig:mini}
\end{figure}

\begin{figure}[t]
\begin{center}
    \includegraphics[width=0.45\textwidth]{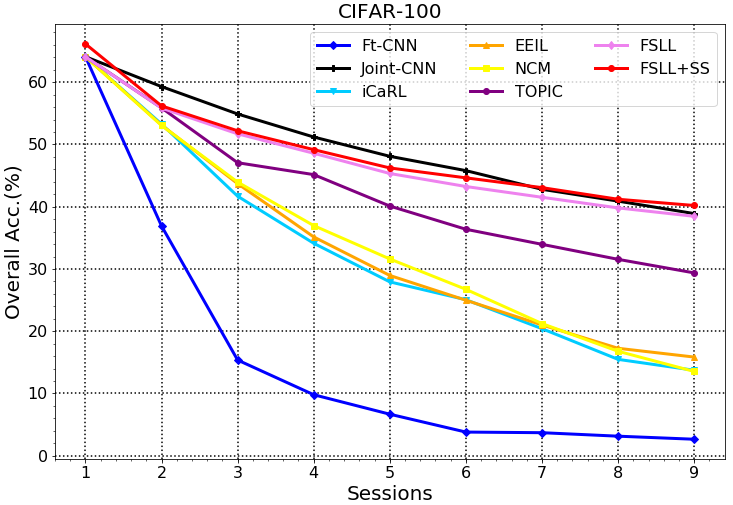}
\end{center}
\caption{Results on CIFAR-100 using the ResNet-18 architecture on the 5-way 5-shot FSCIL setting.}
\label{fig:cifar}
\end{figure}

\subsection{CUB-200 Results}
The results in Table~\ref{tab:CUB-200} indicate that our method significantly outperforms the Ft-CNN model on CUB-200. Our method performs significantly better than the Joint-CNN. This is because CUB-200 contains 100 few-shot classes in this setting and the Joint-CNN model overfits to these classes, resulting in lower overall performance. Our method outperforms the state-of-the-art TOPIC method by an absolute margin of 19.27\%. Even if we exclude the base training set ($D^{(1)}$) accuracy, our method achieves an average accuracy of 27\% on the few-shot training sets. 

While training on the model on $D^{(1)}$, we observed that using an initial learning rate of 0.01 achieves a better session 1 accuracy than reported in \cite{tao2020few}. For completeness, we also provide the results for this model (FSLL*). We perform an additional experiment (FSLL*+SS), where we also train the network on an auxiliary self-supervised rotation prediction task during the training of $D^{(1)}$. 

\subsection{miniImageNet Results}

Fig.~\ref{fig:mini} depicts the performance of different methods on the miniImageNet FSCIL setting. Our method significantly outperforms the Ft-CNN model and performs slightly better than the Joint-CNN model because the Joint-CNN model overfits due to the presence of many few-shot classes. Our method significantly outperforms the state-of-the-art TOPIC model by around 15.07\%. We observe that the performance can be improved further by tuning the weight decay option of the SGD optimizer, which we take as $1e\texttt{-}3$ (FSLL*).  

\subsection{CIFAR-100 Results}

Fig.~\ref{fig:cifar} depicts the performance of different methods on the CIFAR-100 FSCIL setting.  Our method significantly outperforms the state-of-the-art TOPIC model, by an absolute margin of 9.09\%. We perform an additional experiment (FSLL+SS), where we have also trained the network on an auxiliary self-supervised rotation prediction task during the training of $D^{(1)}$.

\section{Ablation Experiments}\label{sec:ablat}
We perform various ablation experiments to validate our method.
\begin{figure}[t]
\begin{center}
    \includegraphics[width=0.4\textwidth]{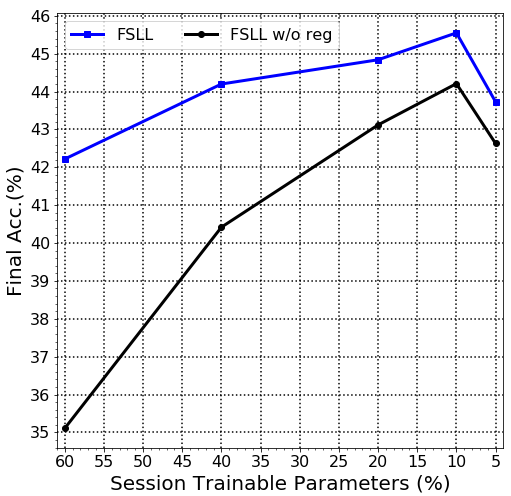}
\end{center}
\caption{Performance of FSLL on the CUB-200 FSCIL setting with and without regularization using different proportions of session trainable parameters.}
\label{fig:regnoreg}
\end{figure}

\subsection{Proportion of Session Trainable Parameters}\label{sec:threshold}
Fig. \ref{fig:regnoreg} shows the effect of increasing the proportion of session trainable parameters on the final accuracy for CUB-200. The final accuracy in Fig. \ref{fig:regnoreg} refers to the session 11 test result ($S_{11}$). If we choose a high threshold for selecting unimportant parameters, then there will be a high proportion of session trainable parameters, and it may include high absolute value/important parameters. We observe that FSLL performance suffers in such a case. FSLL achieves the best performance when the session trainable parameters are 10\% of $\Theta_F$. As we decrease the proportion of session trainable parameters, the proportion of knowledge retention parameters increases, and the performance of the model improves till we reach 10\%. If we choose less than 10\%  of $\Theta_F$ as the session trainable parameters, the model performance starts dropping due to the shortage of trainable parameters (underfitting). 

\subsection{Significance of Regularization}
Fig.~\ref{fig:regnoreg} shows the effect of removing the regularization loss from our method. When the proportion of session trainable parameters is high, the corresponding proportion of knowledge retaining parameters is low, and therefore, the regularization loss plays a critical role in the model performance.  Even when the proportion of session trainable parameters is low ($\approx$10\%), the regularization loss provides improvement to the performance, as shown in Fig.~\ref{fig:regnoreg}.  

\subsection{Choice of Regularization Hyper-Parameter}

\begin{table}[t]
\footnotesize
\centering
\caption{Session 11($S_{11}$) classification results on CUB-200 using the ResNet-18 architecture on the 10-way 5-shot FSCIL setting for different values for the regularization hyper-parameter $\lambda$ values.}
\begin{tabular}{l|ccccc}
\hline
$\lambda$ & 1 & 3 & 5 & 7 & 9\\
\hline
$S_{11}$ & 44.66\% & 44.83\% & \textbf{45.55\%} & 44.96\% & 44.87\%\\

\hline

\end{tabular}
\label{tab:cubablreg}
\end{table}

Table~\ref{tab:cubablreg}, reports the effect of changing the regularization hyper-parameter $\lambda$ on the performance of the model for the CUB-200 dataset in the FSCIL setting. We have reported the session 11 ($S_{11}$) test results in this table. We observe the best model performance for $\lambda=5$, and we use this value of the regularization hyper-parameter for all our experiments.

\subsection{Significance of Cosine Similarity Loss}
We performed ablations to verify the significance of the prototype cosine similarity loss. We observe that in the absence of the prototype cosine similarity loss, the session 11 model performance ($S_{11}$) for the CUB-200 dataset drops from 45.55\% to 44.32\%.

\subsection{Choice of Self-Supervised Auxiliary Task}\label{sec:auxtask}
\begin{table}[t]
\footnotesize
\centering
\caption{Session 11 classification results on CUB-200 using FSLL* for the 10-way 5-shot FSCIL setting for different types of auxiliary self-supervised (SS) task.}
\begin{tabular}{l|cccc}
\hline
Auxiliary SS & Patch & SimClr & Rotation & w/o SS\\
  \hline
$S_{11}$ & 54.56\% & 54.71\% & 55.82\% & 54.21\% \\

\hline

\end{tabular}
\label{tab:cubablrot}
\end{table}
We perform experiments with the auxiliary task as relative patch location prediction \cite{doersch2015unsupervised} (Patch), rotation angle prediction (Rotation) \cite{gidaris2019boosting}  and SimCLR \cite{chen2020simple}. SimCLR utilizes contrastive learning and is the state-of-the-art self-supervision technique. Table \ref{tab:cubablrot}, shows that the rotation-based auxiliary self-supervised task performs significantly better than patch and SimCLR methods.

\subsection{Self-Supervision for Few-Shot Classes}
We also perform experiments to include the self-supervised auxiliary task in the training process for the few-shot training sets ($D^{(t>1)}$) along with the base training set $D^{(1)}$. Our experiments on the CUB-200 dataset show that this results in a session 11 ($S_{11}$) accuracy of 54.34\%, which is lower than 55.82\% achieved by FSLL*+SS. Therefore, using the self-supervised auxiliary task for training on $D^{(t>1)}$ does not produce any benefits.

\section{Conclusion}
We propose a novel Few-Shot Lifelong Learning (FSLL) method for the few-shot class-incremental learning problem. Our method selects very few unimportant parameters as the session trainable parameters to train on every new set of few-shot classes to deal with the problems of overfitting and catastrophic forgetting. We empirically show that FSLL significantly outperforms the state-of-the-art method. We experimentally show that using self-supervision as an auxiliary task can further improve the performance of the model in this setting. 

\bibliography{ref}

\end{document}